\documentclass[conference]{IEEEtran}
\usepackage{cite}

\ifCLASSINFOpdf
\else
\fi
\usepackage[cmex10]{amsmath}
\usepackage{algorithmic}

\usepackage{array}

\usepackage{mdwmath}
\usepackage{mdwtab}

\usepackage{eqparbox}

\usepackage[font=footnotesize]{subfig}
\usepackage{url}

\usepackage{multirow}
\usepackage{makecell}
\usepackage{tabularray}
\usepackage{graphicx}
\usepackage{tikz}
\newlength{\myheight}

\newcommand\copyrighttext{%
\scriptsize
  979-8-3315-1213-2/25/\$31.00~\textcopyright2025 IEEE. Personal use of this material is permitted. Permission from IEEE must be obtained for all other uses, in any current or future media, including reprinting/republishing this material for advertising or promotional purposes, creating new collective works, for resale or redistribution to servers or lists, or reuse of any copyrighted component of this work in other works.}
\newcommand\copyrightnotice{%
\begin{tikzpicture}[remember picture,overlay]
\node[anchor=south,xshift=5,yshift=15pt] at (current page.south) {\parbox{\textwidth}{\copyrighttext}};
\end{tikzpicture}%
}

\begin{document}
%
\title{Trade-offs in Privacy-Preserving Eye Tracking through Iris Obfuscation: A Benchmarking Study}


\author{\IEEEauthorblockN{Mengdi Wang, 
Efe Bozkir, and 
Enkelejda Kasneci 
}
\IEEEauthorblockA{
Technical University of Munich, Munich, Germany \\ 
\{mengdi.wang, efe.bozkir, enkelejda.kasneci\}@tum.de}
}


%


\maketitle
\copyrightnotice

\begin{abstract}
Recent developments in hardware, computer graphics, and AI may soon enable AR/VR head-mounted displays (HMDs) to become everyday devices like smartphones and tablets. Eye trackers within HMDs provide a special opportunity for such setups as it is possible to facilitate gaze-based research and interaction. However, estimating users' gaze information often requires raw eye images and videos that contain iris textures, which are considered a gold standard biometric for user authentication, and this raises privacy concerns. Previous research in the eye-tracking community focused on obfuscating iris textures while keeping utility tasks such as eye segmentation and gaze estimation accurate. Despite these attempts, there is no comprehensive benchmark that evaluates state-of-the-art approaches. Considering all, in this paper, we benchmark blurring, noising, downsampling, rubber sheet model, and iris style transfer to obfuscate users in the data to protect their privacy, and compare their impact on image quality, privacy, utility, and risk of imposter attack on two datasets. We use eye segmentation and gaze estimation as utility tasks, and reduction in iris recognition accuracy as a measure of privacy protection, and false acceptance rate to estimate risk of attack. Our experiments show that canonical image processing methods like blurring and noising cause a marginal impact on deep learning-based tasks. While downsampling, the rubber sheet model, and iris style transfer are effective in hiding user identifiers, iris style transfer, with higher computation cost, outperforms others in both utility tasks, and is more resilient against spoof attacks than the rubber sheet model. Our analyses indicate that there is no universal, optimal approach to balance privacy, utility, and computation burden. Therefore, we recommend practitioners consider the strengths and weaknesses of each approach, and possible combinations of those to reach an optimal privacy-utility trade-off.
\end{abstract}


%
\IEEEpeerreviewmaketitle

\section{Introduction}
Recent advances in hardware, human-computer interaction, and AI will likely lead AR/VR and HMDs to become pervasive, similar to today's smartphones and tablets. Previous works have extensively studied the interaction techniques while using HMDs~\cite{Spittle_etal_2023}, and gaze-based interaction is considered as one of the intuitive ways especially for hands- and voice-free interaction~\cite{plopsky_etal_2022}. The use of gaze interaction and eye-tracking data provides many opportunities for AR and VR; however, it also raises privacy concerns~\cite{bozkir2023eyetrackedVR} considering that it can reveal sensitive user characteristics, including race, sexual preference, or body mass index~\cite{Liebling_and_Preibusch_2014}. Furthermore, it is also possible to authenticate users with iris texture data~\cite{kumar_and_passi_2010} and with eye movements~\cite{katsini_etal_2020}. In fact, as each person's iris texture is unique and as it is treated as visual fingerprints, users can be authenticated with very high accuracy~\cite{daugman_1993}. Such an authentication process is useful when the purpose is user authentication, and data protection is handled responsibly; otherwise, it may give adversaries a good opportunity for impersonation and access to sensitive resources when iris texture data is in hand.

To address this issue, some of the HMD manufacturers do not give access to raw eye image data to application developers and users through their HMDs (e.g., Apple Vision Pro~\cite{avp_2024}, Microsoft HoloLens 2~\cite{hololens2_specs_2023}), others opt for providing access to raw images (e.g., Varjo XR-4~\cite{varjo_xr4_et_2023}). Since gaze estimation techniques mostly utilize eye images to determine the gaze direction, completely limiting access to raw data is not reasonable, especially for computer vision practitioners and researchers. Instead, privacy-preserving methods that degrade iris-based authentication while keeping utility tasks like gaze estimation and eye segmentation accurate, are necessary. Some of the previous works approached this by image blurring~\cite{john_etal_2019, john_etal_2020}, noising~\cite{letitsnow_brendan_etra_2020}, spatial downsampling~\cite{phillips2011impact}, replacing iris region through a rubber sheet model~\cite{Chaudhary_and_Pelz_2020, narkar_and_davidjohn_2024}, and style transfer~\cite{wang2025iris}. However, there is no comprehensive benchmarking study that evaluates the proposed concepts in the literature and compares their privacy-utility trade-offs. We benchmark these concepts using widely recognized HMD datasets, particularly by using the OpenEDS2019~\cite{garbin2019openedsopeneyedataset} and OpenEDS2020~\cite{palmero2020openeds2020openeyesdataset} datasets for privacy protection, semantic segmentation, and gaze estimation tasks. We find that while traditional image processing techniques like noising and blurring do not degrade image utility, they also cannot hide user identity from deep learning-based iris authenticators. Spatial downsampling can notably obfuscate iris identifiable features but also degrades segmentation accuracy for pupil region and performance of the appearance-based estimator. While the rubber sheet model reduces the iris recognition rate to a large degree, it leads to a high risk of iris attack and undermines appearance-based gaze estimation. In comparison, iris style transfer works accurately for utility tasks and significantly reduces iris recognizability without escalating the false acceptance rate, whereas its vanilla pipeline is computationally intensive. In summary, we state that there is no universal best-performing approach to protecting iris in AR/VR settings that harmonizes all aspects including privacy protection, data utility, risk of attack, and computation overhead, and multiple approaches may be combined considering their strengths and weaknesses to find an optimal privacy-utility trade-off. We also publish our code\footnote{\url{https://gitlab.lrz.de/hctl/Iris-Obfuscation-Benchmark}.} for reproducibility.
 

\section{Related work}
With eye tracking becoming ubiquitous with HMDs and smart glasses, despite all the advantages of using it in everyday settings, such as for hands-free interaction, foveated rendering, and understanding human perception and cognition, the use of eye-tracking data comes with privacy risks~\cite{bozkir2023eyetrackedVR}. To mitigate these risks, researchers developed different techniques to protect users' privacy, with differential privacy~\cite{diffprivacy_bozkir_2021, Steil_etal_2019}, secure multi-party computation~\cite{ozdel_etal_2024}, and signal processing-based approaches including spatial and temporal downsampling~\cite{9382914}. Yet, these approaches focus on the protection of eye movements and their features after gaze estimation takes place, considering eye movements and features can reveal sensitive information about users~\cite{Liebling_and_Preibusch_2014, Kroger2020}. However, when the raw eye data is considered, one of the most sensitive data types is iris texture, as it can uniquely identify individuals.  
To protect iris information, previous research has frequently focused on obscuring iris texture to impede iris-based user identification~\cite{Eskildsen_etal_2021}. John et al.~\cite{john_etal_2019} proposed Gaussian blurring to achieve this while retaining gaze estimation capability. Further, John et al.~\cite{john_etal_2020} showed that with such a blurring mechanism to approach optical defocus in a conversational setting in VR, it is possible to find a trade-off that eye movement changes due to the privacy protection can go unnoticed. In another work~\cite{letitsnow_brendan_etra_2020}, the authors introduced pixel noise on iris textures and stated similarly that gaze estimation errors due to privacy protection are in acceptable ranges. Chaudhary and Pelz~\cite{Chaudhary_and_Pelz_2020} proposed to replace iris segments of an eye image with fake irises using a rubber sheet model and showed that such manipulated eye images can still be useful for eye semantic segmentation while protecting individuals' privacy. Later, the rubber sheet method has been studied in the opposite direction, namely constructing an iris attack by swapping in attackers' iris texture to fool liveness detection and iris authentication systems~\cite{narkar_and_davidjohn_2024}. Recently, Wang et al.~\cite{wang2025iris} proposed utilizing style features for iris recognition and suggested protecting privacy via iris style transfer accordingly. The authors showed their method maintains eye image utility for eye segmentation and for gaze estimation using both model- and appearance-based methods, while suppressing the risk of impersonation attack. 

Despite the importance of preserving eye image privacy, prior works that systematically compare different iris obfuscation methods~\cite{Eskildsen_etal_2021} are commonly limited to primitive image processing techniques like filtering and noising and their influence on traditional iris recognition systems and utility tasks, which commonly rely on handcrafted kernels and objectives~\cite{wildes1997iris, ali2007recognition, daugman2009iris}. A comprehensive benchmarking study in the literature that compares the state-of-the-art methods for iris obfuscation is missing. We address this in our work by comparing different iris manipulation methods regarding their effect on image quality, privacy, risk of attack, and utility, using deep learning-based models on two publicly available datasets collected in VR environments. 

\section{Experiments}
In this study, we extensively benchmarked five iris obfuscation methods, including blurring, noising, downsampling, rubber sheet model, and iris style transfer, regarding their impact on image quality, privacy, and utility. Particularly, the influence on image privacy involves the impact on iris recognition and the risk of malicious use, while the study on data utility covers the performance of eye semantic segmentation and gaze estimation using obfuscated eye images.

\subsection{Methods}
\setlength{\myheight}{2cm}
\begin{figure*}[!ht]
    \centering
    \subfloat[raw iris]{\includegraphics[height=\myheight, keepaspectratio]{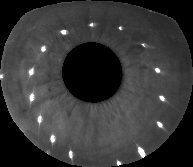} 
    \label{fig:raw}}
    \subfloat[attacker iris]{\includegraphics[height=\myheight, keepaspectratio]{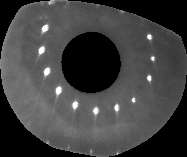}
    \label{fig:attacker}}
    \subfloat[gaussian blurring]{\includegraphics[height=\myheight, keepaspectratio]{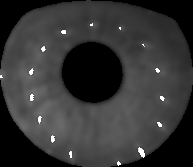}
    \label{fig:blur}}
    \subfloat[gaussian noising]{\includegraphics[height=\myheight, keepaspectratio]{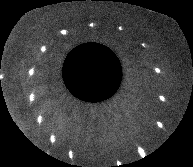}
    \label{fig:noise}}
    \subfloat[downsampling]{\includegraphics[height=\myheight, keepaspectratio]{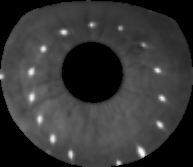}
    \label{fig:dowmsample}}
    \subfloat[rubber sheet]{\includegraphics[height=\myheight, keepaspectratio]{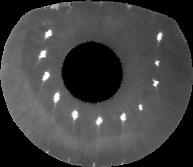}
    \label{fig:rubber}}
    \subfloat[iris style transfer]{\includegraphics[height=\myheight, keepaspectratio]{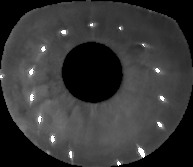}
    \label{fig:ist}}
    \caption{Examples of raw irises (a and b) and obfuscated irises (c - g).}
    \label{fig:manipulated}
\end{figure*}

\begin{table*}[!ht]
    \caption{Impact of iris obfuscation methods on privacy, utility, and risk of attack. Baseline: no obfuscation. Metrics: classification accuracy for iris recognition, false acceptance rate (FAR) for attack success, intersection over union (IoU) for eye segmentation, and degree error for gaze estimation. $s$: scaling factor. $e$: style transfer iteration. c1: feed-forward CNN feature-driven classifier. c2: style feature-driven classifier. m1: model-based gaze estimator. m2: appearance-based gaze estimator.}
    \label{tab:results}
    \centering
    \tiny
    \resizebox{\textwidth}{!}{%
    \begin{tabular}{l | l | c c | c c | c c c c | c c}
        Method & Parameter & \multicolumn{2}{c|}{Iris Recognition} & \multicolumn{2}{c|}{Attack Success} & \multicolumn{4}{c|}{Eye Segmentation} & \multicolumn{2}{c}{Gaze Estimation} \\
        & & c1 & c2 & c1 & c2 & skin & sclera & iris & pupil & m1 & m2 \\
        \hline
        Baseline & NA & 0.962 & 0.978 & 0.001 & 0.000 & 0.995 & 0.932 & 0.956 & 0.945 & 6.970° & 3.190° \\
        \hline
        \multirow{5}{*}{\makecell[l]{Blurring}} 
        & $\sigma = 1$ & 0.959 & 0.964 & 0.001 & 0.000 & 0.995 & 0.932 & 0.956 & 0.945 & 6.970° & 3.193° \\
        & $\sigma = 2$ & 0.953 & 0.963 & 0.001 & 0.000 & 0.995 & 0.932 & 0.956 & 0.944 & 6.971° & 3.197° \\
        & $\sigma = 3$ & 0.951 & 0.963 & 0.001 & 0.000 & 0.995 & 0.932 & 0.956 & 0.944 & 6.970° & 3.198° \\
        & $\sigma = 4$ & 0.951 & 0.963 & 0.001 & 0.000 & 0.995 & 0.932 & 0.956 & 0.944 & 6.971° & 3.199° \\
        & $\sigma = 5$ & 0.949 & 0.963 & 0.001 & 0.001 & 0.995 & 0.932 & 0.956 & 0.944 & 6.970° & 3.199° \\
        \hline
        \multirow{5}{*}{\makecell[l]{Noising}} 
        & $\sigma = 0.01$ & 0.961 & 0.975 & 0.001 & 0.000 & 0.995 & 0.932 & 0.956 & 0.945 & 6.970° & 3.189° \\
        & $\sigma = 0.05$ & 0.950 & 0.964 & 0.001 & 0.002 & 0.995 & 0.932 & 0.956 & 0.945 & 6.970° & 3.192° \\
        & $\sigma = 0.1$ & 0.949 & 0.963 & 0.002 & 0.002 & 0.995 & 0.932 & 0.955 & 0.942 & 6.970° & 3.199° \\
        & $\sigma = 0.2$ & 0.948 & 0.963 & 0.002 & 0.002 & 0.995 & 0.931 & 0.953 & 0.936 & 6.975 ° & 3.217° \\
        & $\sigma = 0.5$ & 0.948 & 0.963 & 0.001 & 0.000 & 0.995 & 0.929 & 0.947 & 0.931 & 7.019° & 3.259° \\
        \hline
        \multirow{5}{*}{\makecell[l]{Down- \\ sampling}}
        & $s = 1.5$ & 0.831 & 0.222 & 0.001 & 0.006 & 0.995 & 0.925 & 0.944 & 0.917 & 7.039° & 3.425° \\
        & $s = 2$ & 0.662 & 0.150 & 0.004 & 0.007 & 0.995 & 0.924 & 0.940 & 0.903 & 7.042° & 3.872° \\
        & $s = 3$ & 0.462 & 0.070 & 0.006 & 0.004 & 0.994 & 0.923 & 0.935 & 0.883 & 7.053° & 4.353° \\
        & $s = 4$ & 0.344 & 0.031 & 0.008 & 0.006 & 0.994 & 0.922 & 0.932 & 0.865 & 7.080° & 4.854° \\
        & $s = 5$ & 0.278 & 0.026 & 0.009 & 0.005 & 0.994 & 0.922 & 0.930 & 0.850 & 7.108° & 5.287° \\
        \hline
        \makecell[l]{Rubber Sheet} & NA & 0.069 & 0.025 & 0.245 & 0.369 & 0.994 & 0.918 & 0.931 & 0.897 & 7.008° & 19.574° \\
        \hline
        \multirow{5}{*}{\makecell[l]{Iris Style \\ Transfer}}
        & $e = 1$ & 0.835 & 0.175 & 0.005 & 0.041 & 0.995 & 0.928 & 0.949 & 0.933 & 7.016° & 3.187° \\
        & $e = 10$ & 0.835 & 0.175 & 0.005 & 0.041 & 0.995 & 0.928 & 0.949 & 0.933 & 7.016° & 3.187° \\
        & $e = 50$ & 0.773 & 0.112 & 0.008 & 0.054 & 0.995 & 0.927 & 0.948 & 0.930 & 6.976° & 3.199° \\
        & $e = 100$ & 0.742 & 0.093 & 0.011 & 0.054 & 0.995 & 0.927 & 0.947 & 0.928 & 6.988° & 3.214° \\
        & $e = 200$ & 0.696 & 0.064 & 0.012 & 0.062 & 0.995 & 0.925 & 0.945 & 0.924 & 7.026° & 3.329° \\
    \end{tabular}
    }
\end{table*}

\subsubsection{Blurring}
Blurring, which is a specific type of filtering, is a classical image processing technique that is often used for noise reduction and edge softening. It operates by convolving the image with a kernel function $K$.
Given an image $I(x, y)$, the image convolution can be represented as:
\begin{equation}
I_{\text{blurred}}(x, y) = \sum_{i=-k}^{k} \sum_{j=-k}^{k} K(i, j) I(x-i, y-j)
\end{equation}
where $k$ denotes kernel size. Blurring is also a low-pass filter. Phillips and Komogortsev~\cite{phillips2011impact} investigated the influence of Gaussian blurring on iris identification. Other work also hypothesized that eye images consist of low-frequency eye-tracking signals and high-frequency identifiable iris features, and used Gaussian blurring to preserve privacy~\cite{john_etal_2019, john_etal_2020}. 

\subsubsection{Noising}
Image noising refers to the process of deliberately adding noise to an image. It can be modeled by:
\begin{equation}
I_{\text{noisy}}(x, y) = I(x, y) + N(x, y), N(x, y) \sim P_N(\theta) 
\end{equation}
with $N(x, y)$ being random noise sampled from a specific distribution $P_N(\theta)$ parametrized by $\theta$. Noising can be used as a privacy-enhancing tool and is often employed in privacy mechanisms like differential privacy. Previous work proposed to intelligently add noise to specific image regions to protect privacy while preserving image aesthetics~\cite{rana2023novel}. John et al.~\cite{letitsnow_brendan_etra_2020} introduced pixel-wise noise to eye images to obscure iris signature and achieved $(0, \delta)$-differential privacy.

\subsubsection{Downsampling}
Image downsampling is the process of reducing the resolution of an image by decreasing the number of pixels while trying to retain visual information. For an image $I(x, y)$ and a scaling factor $s$, the reduced image can be represented as $I_{\text{down}}(x', y') = I(sx', sy')$. Since the decreased image resolution can distort sensitive information in the image, it can also be used to preserve privacy. For instance, Phillips and Komogortsev~\cite{phillips2011impact} showed that the resolution of eye images has a strong impact on iris identification.

\subsubsection{Rubber Sheet}
The rubber sheet model is an image warping and transformation technique. In iris recognition systems, the rubber sheet model can be used in combination with polar transformation to preprocess iris texture~\cite{daugman2009iris}. Recently, the rubber sheet model has been studied as a privacy-preservation method by swapping the real iris texture with a fake template~\cite{Chaudhary_and_Pelz_2020}. Later, the same method was applied in the opposite direction, where the researchers switched the victims' iris texture with attackers' to fool liveness detection and iris authentication systems~\cite{narkar_and_davidjohn_2024}.

\subsubsection{Iris style transfer}
Iris style transfer~\cite{wang2025iris} is a recently proposed eye privacy-enhancing technique inspired by neural style transfer~\cite{gatys2015neural, gatys2016image}, which is a non-photorealistic image synthesis method~\cite{gooch2001non, strothotte2002non}. It works under the premise that images consist of content and style components, which can be separated and manipulated independently. By iteratively optimizing the content and style objectives, neural style transfer recombines the content and style of different images. Li et al.~\cite{li2017demystifying} proved that neural style transfer can be modeled as a domain adaptation problem in feature space, and proposed a novel way to compute style loss as statistical matching:
\begin{equation}
    \mathcal{L}^l_{style} = \frac{1}{N^l} \sum_{i=1}^{N^l} \left( (\mu^i_{F^l_S} - \mu^i_{F^l_I} )^2 +(\sigma^i_{F^l_S} - \sigma^i_{F^l_I})^2 \right)
\end{equation}
where $F^l_S$ and $F^l_I$ are, respectively, feature maps of the style and target images extracted by a convolution neural network (CNN) at layer $l$, $\mu$ and $\sigma$ represent the mean and standard deviation of the feature maps, and $N^l$ is number of channel in the feature maps. Wang et al.~\cite{wang2025iris} proposed style feature-based iris recognition, which is robust against image rotation and perspective shift, and suggested transferring iris texture style to block iris recognition accordingly. Unlike the rubber sheet model which can be used for adversarial purposes~\cite{Chaudhary_and_Pelz_2020, narkar_and_davidjohn_2024}, the risk of malicious use of iris style transfer is low.

\subsection{Datasets, models, and tasks}
We employed two publicly available datasets in our experiments, namely OpenEDS2019~\cite{garbin2019openedsopeneyedataset} and OpenEDS2020~\cite{palmero2020openeds2020openeyesdataset}, collected from VR HMDs with near-eye cameras. The former is from 152 users for eye semantic segmentation, whereas the latter is specialized for gaze estimation. We investigated the influence of iris obfuscation methods on gaze estimation using OpenEDS2020, while we conducted the benchmarks regarding data privacy and eye segmentation on OpenEDS2019.

To explore the influence of iris obfuscation methods, we compared the outcome of different tasks before and after manipulation. To approach iris obfuscation, the iris regions were first segmented out from eye images. We utilized two publicly available pre-trained segmentation networks for this purpose, i.e., RITnet~\cite{chaudhary2019ritnet} and EfficientNet~\cite{openeds2020_seg_model}, for OpenEDS2019 and OpenEDS2020, respectively. Then, the glint blobs were detected with thresholding. Later, the iris regions went through different obfuscation techniques, and the glints were added back. In the end, the manipulated irises were placed back onto the eye images for post-obfuscation evaluation. 

We emulated iris recognition with iris-based user classification among 152 users, and trained two classifiers. The first one uses the common CNN feed-forward features extracted by VGG19~\cite{simonyan2014very} network, whilst the second one uses the style features~\cite{wang2025iris}. For the impact on eye segmentation, we used the RITnet and EfficientNet models again. To examine the influence on gaze estimation, we trained both model- and appearance-based gaze estimators~\cite{hansen2009eye}. The former relies on eye landmark features, while the latter utilizes a ResNet50~\cite{he2016deep} to encode hidden representations. After being trained or loaded, all models were frozen for benchmarks.

Particularly, the rubber sheet model and iris style transfer may be used for malicious purposes, i.e., impersonation attacks by swapping iris texture or transferring iris style of an attacker's eye image into a victim's. To simulate such a scenario, for each eye image of user A, we randomly sampled another eye image of another user B, and swapped the iris texture or transferred the iris style of B's image into A's. The relevant measurement is FAR (false acceptance rate), a key metric for biometric authentication systems, and is computed as the ratio between the number of unauthorized but falsely accepted users and the total number of spoofing attacks.

In our benchmarks, we used Gaussian kernel for image blurring with $\sigma$ in $\{1, 2, 3, 4, 5\}$ and kernel size $= 6\sigma + 1$, and Gaussian noise for image noising with $\mu = 0$ and $\sigma$ from $\{0.01, 0.05, 0.1, 0.2, 0.5\}$. For downsampling, we set the interpolation strategy to the nearest neighbor and explored scaling factors from 1.5 to 5. 
For iris style transfer, we fixed the ratio between content and style objective weights to 1, and performed optimization for $\{1, 10, 50, 100, 200\}$ iterations.

\subsection{Evaluation}
We first analyzed the obscured iris images, as depicted in Figure~\ref{fig:manipulated}. Visually, blurring ($\sigma = 2$) and downsampling (scaling factor 2) resulted in a similar distortion effect, while the latter also blurred the boundaries of glint blobs. In contrast, noising introduced noticeable white noise even with a relatively small variance ($\sigma = 0.05$). Different from the three computation-friendly methods that strongly degraded image quality, the rubber sheet model and iris style transfer (200 iters) generated images with decent quality and realism. While the rubber sheet model successfully deformed the attacker's iris according to the victim's iris shape, it also involved the glints, which could affect gaze estimation approaches that rely on them. Iris style transfer brought in subtle but sensible changes in iris patterns and maintained image visualization to a large extent.

The quantitative results, measured through iris recognition accuracy, FAR, eye segmentation accuracy, and gaze estimation error, were reported in Table~\ref{tab:results}. We noticed that while blurring and noising caused negligible drop in image utility across different variance magnitudes, they did not particularly protect user identity, as iris identification using the manipulated iris images preserved its accuracy. This contradicts the previous findings in~\cite{Eskildsen_etal_2021}, where the impact of noising and blurring was remarkable. We believe the main reason for such discrepancy lies in that in~\cite{Eskildsen_etal_2021} the tasks were mainly conducted with canonical image processing methods that commonly rely on handcrafted kernels, while our models are pre-trained CNNs, which are more robust against noise and distortion. Another lightweight manipulation technique, namely image downsampling, with increasing scaling factor, resulted in a remarkable decrease in recognition accuracy from 96.2\% to 27.8\% and from 97.8\% to 2.6\%, albeit sharing a similar visual effect with blurring. In the meanwhile, downsampling noticeably deteriorated segmentation accuracy for the pupil region, whilst other eye regions were slightly impacted. While downsampling caused a subtle influence on the model-based gaze estimation approach, it resulted in a notable increase in prediction error for the appearance-based estimator. This can be attributed to that the model-based method necessitates eye segmentation as a preprocessing step, and since downsampling did not lower segmentation accuracy much, the model-based gaze estimator maintained its performance considerably.

For the iris swapping through the rubber sheet model, we observed a substantial drop in iris recognition accuracy from 96.2\% to 6.9\% for the CNN feature-based classifier and from 97.8\% to 2.5\% for the style feature-based identifier, indicating that rubber sheet model can hinder user identification. However, it also escalated the risk of impersonation attack, as FAR increased from 0.1\% to 24.5\% and from 0\% to 36.9\% in our two settings. This is in accordance with our expectation since the rubber sheet model substitutes the attacker's iris for the victim's. On the utility side, the rubber sheet model maintained the eye segmentation accuracy in essence, as it reshapes the attacker's iris according to the shape of the victim's iris. The model-based gaze estimation performance was preserved as well for this reason. On the contrary, as expected, we noticed a prominent degradation in appearance-based gaze estimation performance, indicated by an escalation of gaze error from 3.190° to 19.574°, because the whole iris region in the manipulated eye image belonged to someone else. The other computationally intensive method, namely iris style transfer, successively reduced the iris recognition accuracy of the style-based classifier to 17.5\% - 6.4\% with increasing iterations, while causing a lighter but noticeable impact on the CNN feature-based authenticator. Different from rubber sheet model which is prone to spoof attacks, iris style transfer merely caused a negligible growth in attack success chance, and the utility tasks' performance was preserved to a large extent.

From the perspective of computation overhead, primitive image processing methods, including blurring, noising, and scaling, are lightweight and can be performed easily in real time. Iris swapping using a rubber sheet model involves polar transformation and row-wise interpolation, and is hence more computationally demanding. Compared to other benchmarked methods, the vanilla style transfer operates through iterative optimization~\cite{jing2019neural} and is thus the most resource-intensive. For a more efficient eye image and even video stylization in real time, advanced variants~\cite{johnson2016perceptual, huang2017arbitrary, gupta2017characterizing} should be applied.

\section{Conclusion}
In this paper, we conducted extensive benchmarks of five iris obfuscation methods, including blurring, noising, downsampling, rubber sheet model, and iris style transfer, regarding their impact on eye image quality, privacy, utility, and risk of attack, in deep learning-based tasks. Our results indicate that there is no globally superior approach that balances all factors including privacy, utility maintenance, resilience against impersonation attacks, and computation overhead. Therefore, we suggest that the choice of iris manipulation method should be determined according to individual scenarios, with consideration to potential combinations of methods.


\section*{Acknowledgment}
We acknowledge the funding by the Deutsche Forschungsgemeinschaft (DFG) – Project number KA 4539/5-1.



\bibliographystyle{IEEEtran}
\clearpage
\bibliography{refs}
%
%
%

\end{document}